\documentclass[lettersize,journal]{IEEEtran}
\usepackage{amsmath,amsfonts}
\usepackage{algorithmic}
\usepackage{algorithm}
\usepackage{array}
\usepackage[caption=false,font=normalsize,labelfont=sf,textfont=sf]{subfig}
\usepackage{textcomp}
\usepackage{stfloats}
\usepackage{url}
\usepackage{verbatim}
\usepackage{graphicx}
\usepackage{cite}
\usepackage{multirow}
\usepackage{placeins}

\graphicspath{ {images/} }

\begin{document}


\title{\vspace{-0.4cm} RemoteNet: Remote Sensing Image Segmentation Network based on Global-Local Information \vspace{-0.1cm}}


\author{Satyawant Kumar, Abhishek Kumar, Dong-Gyu Lee \vspace{-1.0cm}
\thanks{(Corresponding author: dglee@knu.ac.kr)}}

\markboth{Journal of \LaTeX\ Class Files,~Vol.~14, No.~8, August~2021}%
{Shell \MakeLowercase{\textit{et al.}}: A Sample Article Using IEEEtran.cls for IEEE Journals}


\maketitle

\begin{abstract}
Remotely captured images possess an immense scale and object appearance variability due to the complex scene. It becomes challenging to capture the underlying attributes in the global and local context for their segmentation. Existing networks struggle to capture the inherent features due to the cluttered background. To address these issues, we propose a remote sensing image segmentation network, RemoteNet, for semantic segmentation of remote sensing images. We capture the global and local features by leveraging the benefits of the transformer and convolution mechanisms. RemoteNet is an encoder-decoder design that uses multi-scale features. We construct an attention map module to generate channel-wise attention scores for fusing these features. We construct a global-local transformer block (GLTB) in the decoder network to support learning robust representations during a decoding phase.
Further, we designed a feature refinement module to refine the fused output of the shallow stage encoder feature and the deepest GLTB feature of the decoder. Experimental findings on the two public datasets show the effectiveness of the proposed RemoteNet.
\end{abstract}




\vspace{-4mm}
\section{Introduction}
\label{sec:sample1}

\IEEEPARstart{A}{erial} or remote sensing images contain massive details, which can be advantageous for scene understanding, traffic estimation, and infrastructure planning. Recently, semantic segmentation of remote sensing imagery has shown remarkable results in various contexts \cite{osco2021review} and has become a new research option. Remotely fetched images generally possess complex scenery with numerous scene and outlook variations. Hence, it often poses a challenge for its segmentation. Methods based on convolutional neural networks (CNNs) have always dominated the segmentation tasks. Existing methods \cite{li2021multiattention, wang2022unetformer, su2022semantic, wang2021transformer, kirillov2019panoptic} employ different encoder variants based on fully convolutional network (FCN) \cite{long2015fully}. However, FCN-based backbones show inaccurate predictions \cite{zheng2021rethinking}, and the receptive field becomes constant after a specific boundary. Also, the receptive fields are local due to small kernels in the CNN, allowing it to extract the fine-grained local details \cite{yuan2021incorporating}. However, it struggles to capture global contextual details \cite{strudel2021segmenter, zheng2021rethinking}.

\begin{figure}[t]
\centering
\includegraphics[width=1.0\linewidth]{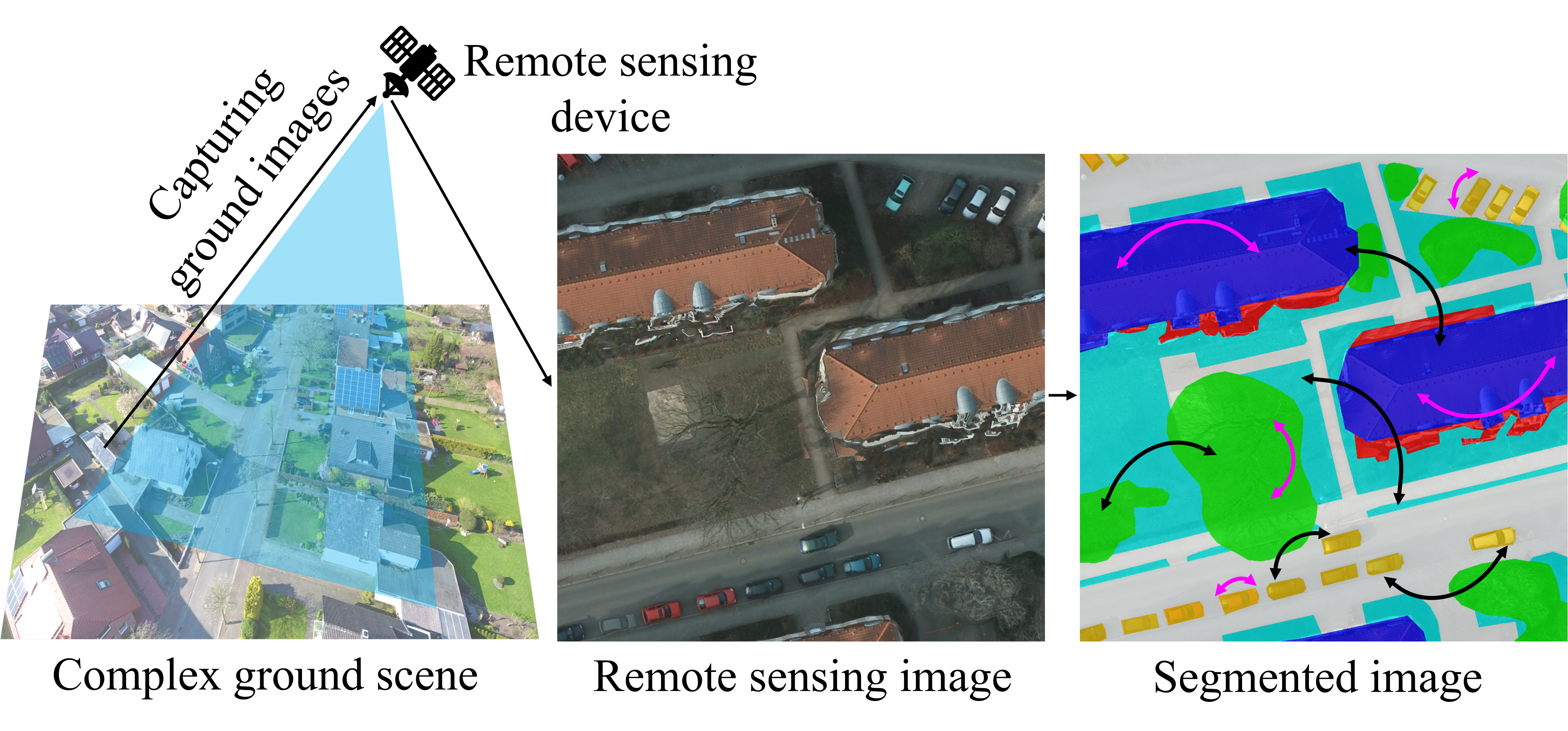}
\caption{The illustration of global and local context details in semantic segmentation of remote sensing images. The black and pink arrows in the segmented image illustrate the global and local context details, respectively. \vspace{-8mm}}
\label{1_intro}
\end{figure}

Capturing global and local contexts is crucial \cite{osco2021review, wang2022unetformer} for segmenting extremely complex images. The black and pink arrows in Fig. \ref{1_intro} illustrate the global and local context details in the segmented image, respectively. Adjacent pixels of the same class represent the local context, whereas same-class pixels far from each other represent the global context. Existing networks \cite{li2021multiattention, wang2022unetformer, ma2021factseg, wang2021transformer, sun2023target, niu2021hybrid, kumar2022semantic} explicitly created for remote sensing scenes yield acceptable results, but they struggle to capture the actual intrinsic properties of the images. The majority of the networks adopt a ResNet-based feature extractor. However, CNNs alone limit their potential in highly cluttered domains due to the fixed receptive views. Recent researches \cite{wang2022unetformer, wang2021transformer, wang2022novel, kumar2022semantic} has tried integrating CNNs with the transformers to improve performance. Transformers can potentially capture long-range relations in an image \cite{zheng2021rethinking}. However, they use transformer modules only in a particular section of the entire network. They show some improvement in the results but still need more precise results, and capturing the contextual details remains challenging.

In this work, we propose a RemoteNet, an encoder-decoder framework for precise semantic segmentation of remote sensing images. It combines the advantage of the transformers and CNNs in capturing the global and fine-grained local contextual details, respectively. It uses multi-scale features to capture details of varying scales in the complex scene. The features generated by the encoder are aggregated with their corresponding decoder features using a feature fusion module. We construct an attention map module (AMM) to generate channel-wise attention scores for the feature fusion. Fusing the encoder-decoder attributes makes the absolute representation semantically rich, which helps achieve precise results. We construct a global-local transformer block (GLTB) to design our decoder network on top of the transformer-based encoder network. Finally, we construct a feature refinement module (FRM) to refine the fused output of the shallow stage encoder network feature and the deep GLTB feature of the decoder network.

In summary, our work has made the following contributions:
\begin{enumerate}
  \item We propose RemoteNet to segment remote sensing images and capture the intrinsic details in complex scenes.
  \item We construct a fusion module using the AMM for fusing the multi-scale encoder and decoder features.
  \item We construct the GLTB to retain rich context details in the decoder network.
  \item We construct the FRM to refine the deepest decoder network feature before segmentation.
  \item Experiments on the LoveDA and Potsdam datasets reveal that RemoteNet performs competitively versus state-of-the-art methods. 
\end{enumerate}

\section{Methodology}
\begin{figure}[h]
\centering
\includegraphics[width=1.0\linewidth]{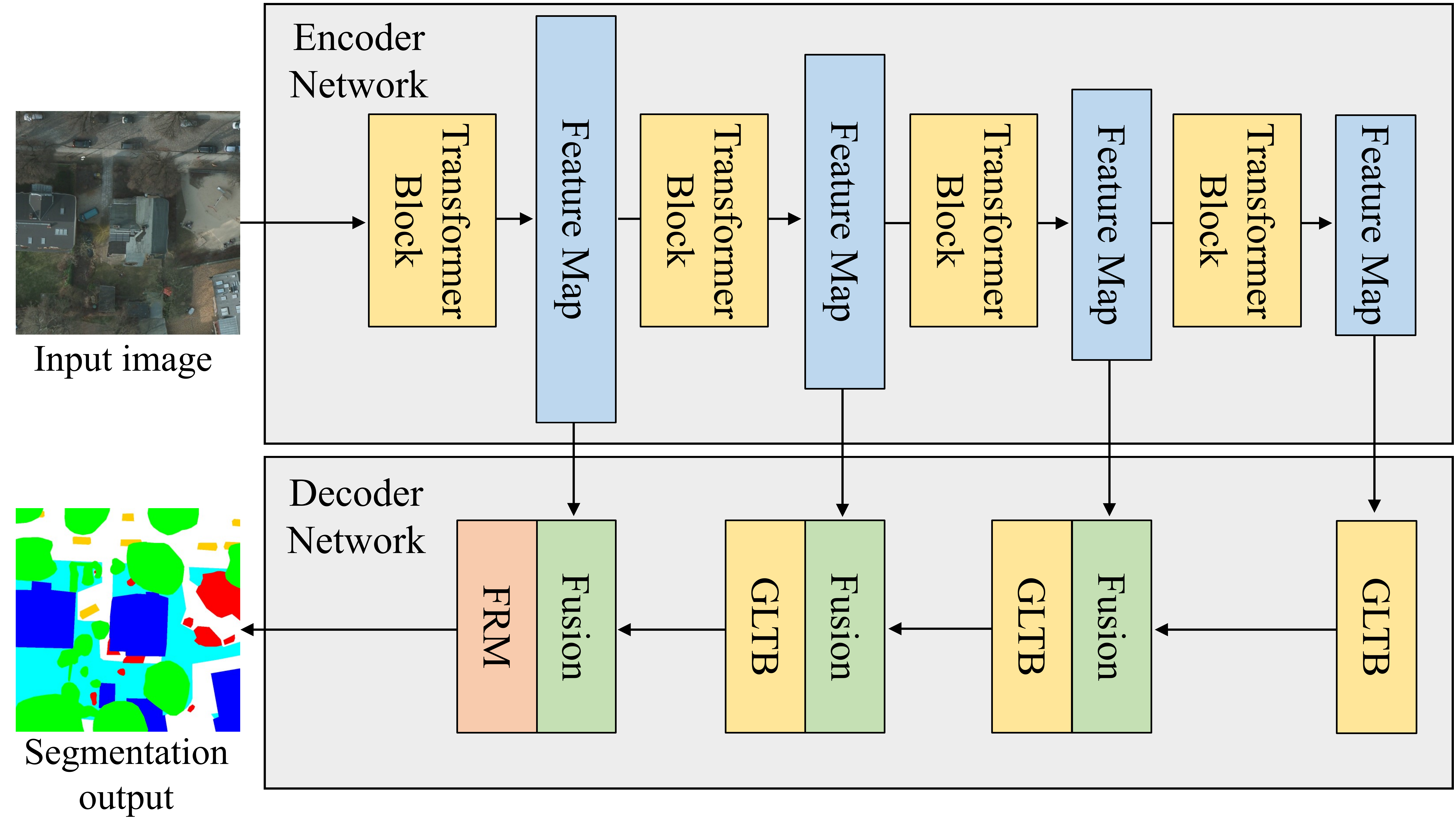}
\caption{The schematic representation of the proposed encoder-decoder based remote sensing image segmentation network, RemoteNet.}
\label{2_overall_framework}
\end{figure}

This section discusses the design of the proposed RemoteNet. Fig. \ref{2_overall_framework} presents its overview, the encoder network generates multi-scale features, and the decoder network processes them for the segmentation result. \vspace{-3mm}

\vspace{-7.5mm}
\subsection{Encoder network}
\begin{figure}[h]
\centering
\includegraphics[width=1.0\linewidth]{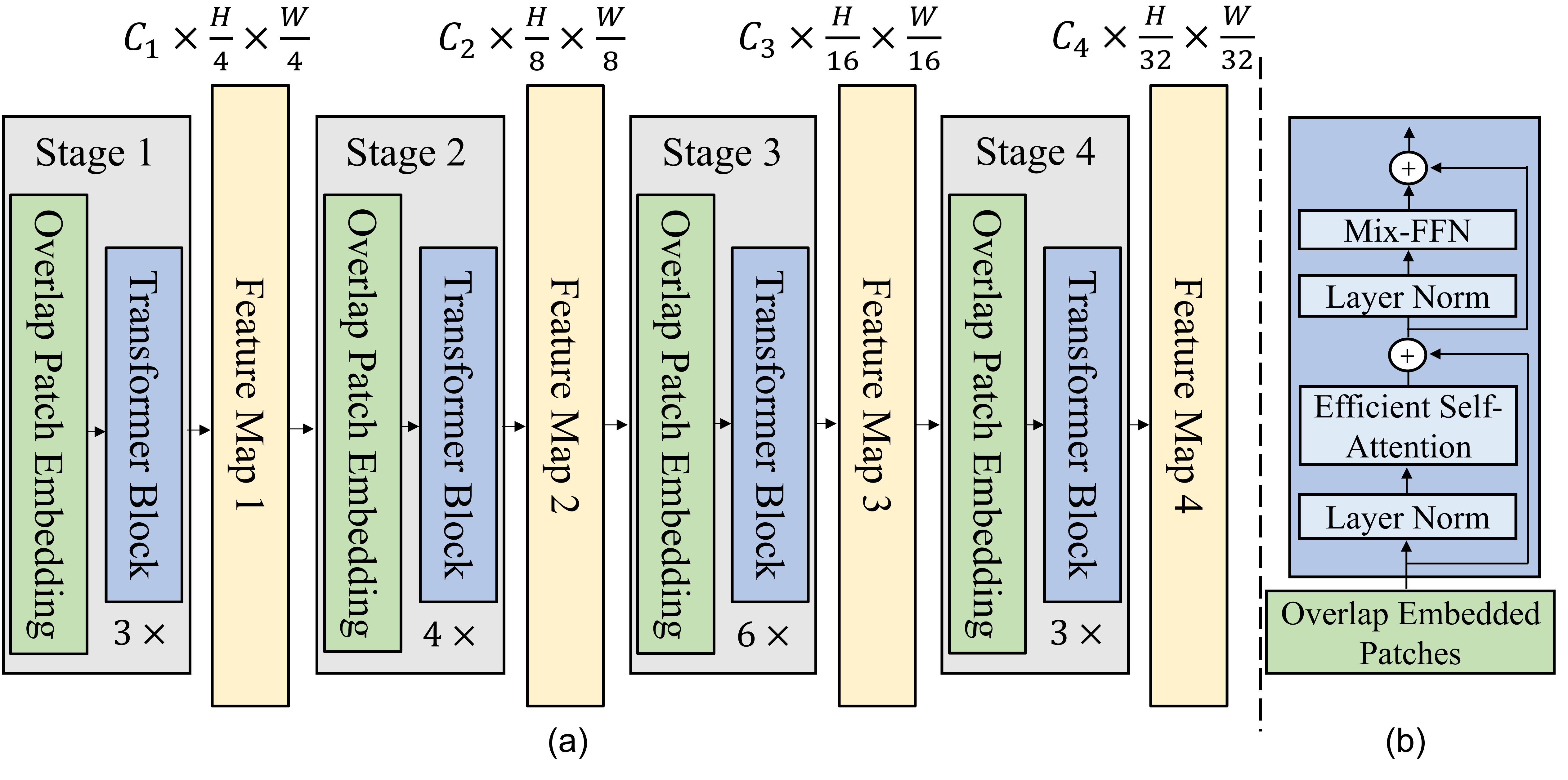}
\caption{(a) The illustration of the encoder network. It consists of four stages and generates multi-scale features. (b) Transformer block module.}
\label{3_encoder_network}
\end{figure}

The encoder contains four stages, where each stage comprises an overlap patch embedding and transformer block modules \cite{xie2021segformer} as shown in Fig. \ref{3_encoder_network}$\left(a \right)$. \vspace{-3mm}

\vspace{-2mm}
\subsubsection{Overlap Patch Embedding (OPE)}
It performs an overlapped tokenization of an input image or the feature maps from each stage and generates multi-scale features. Hence, the OPE fuses the local details among neighboring pixels. The adjacent pixels are very much correlated, and fusing its low-level details helps capture the local context attribute \cite{yuan2021incorporating}.

\subsubsection{Transformer Block}
\label{transformer_encoder}
It contains efficient self-attention (SA) and mix feed forward network (Mix-FFN) blocks, as described in Fig. \ref{3_encoder_network}$\left(b \right)$.
Multiple SA operation is performed to generate multi-head self-attention. We perform a SA as follows: \vspace{-3mm}
\begin{equation}
Self Attention(Q, K, V) = Softmax(\frac{QK^{T}}{\sqrt{C}})V,
\end{equation}
where the $Q$, $K$, and $V$ stand for query, key, and value vectors, respectively.
Since SA captures the long-range attributes \cite{zheng2021rethinking, yuan2021incorporating}, it assists in sustaining the prosperous global context details.
Parallel to the SA operation, we append a 2D positional attention module \cite{li20222dsegformer} to preserve the relationship among the tokens based on their 2D distances.
Then the Mix-FFN processes the SA output, which consists of two linear layers connected by nonlinearity and a depth-wise convolution. Convolutions have a local receptive field and can extract local attributes about adjacent pixels \cite{yuan2021incorporating, xie2021segformer}.

\vspace{-4mm}
\subsection{Decoder network} \vspace{-1.mm}
The decoder network contains GLTB, feature fusion, and FRM, as shown in Figure \ref{2_overall_framework}. The whole decoder uses a channel dimension of $64$.

\subsubsection{Global-Local Transformer Block (GLTB)}
\begin{figure}[h]
\centering
\includegraphics[width=1.0\linewidth]{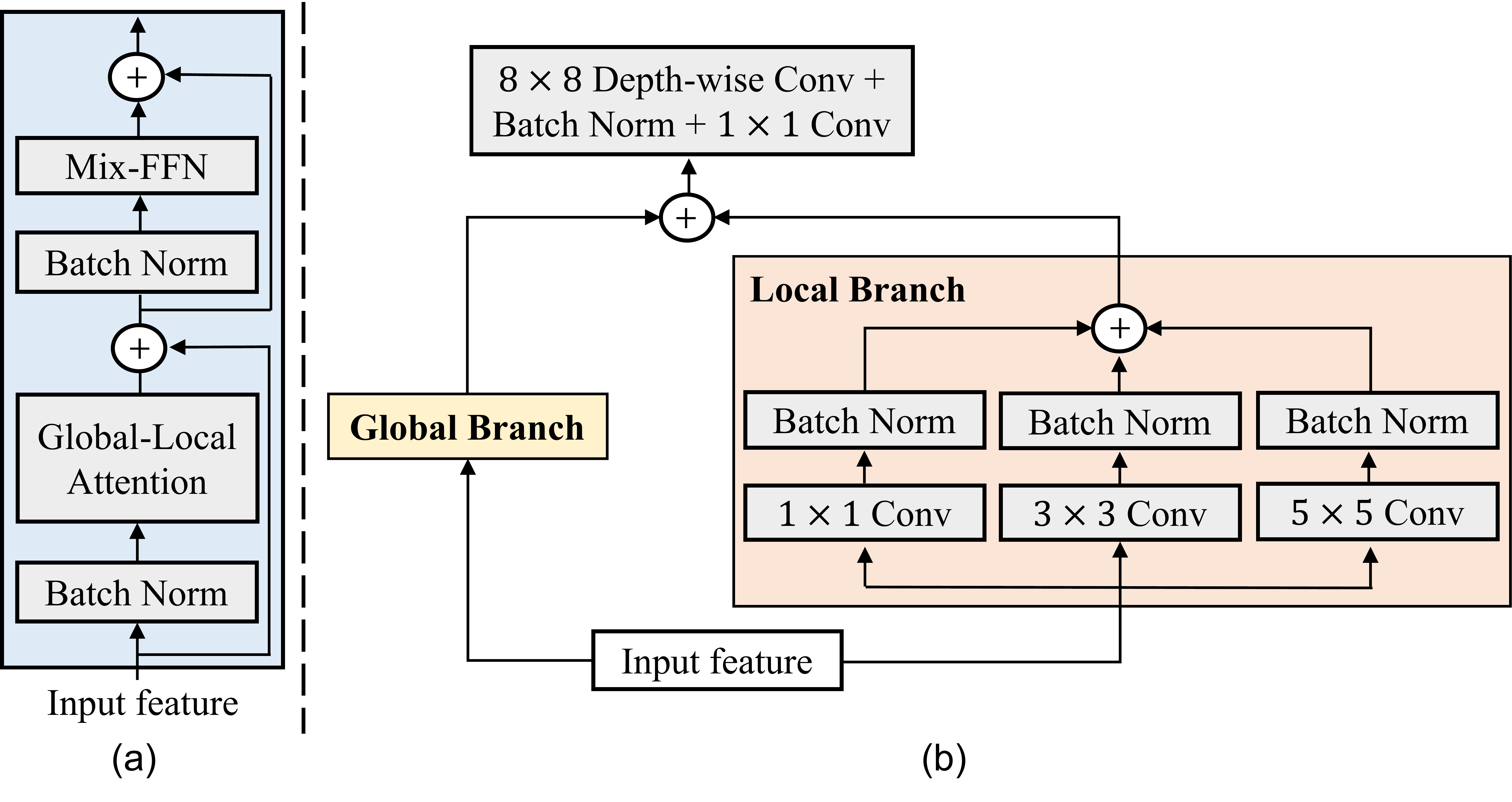}
\caption{(a) The illustration of global-local transformer block. (b) Global-local attention module. \vspace{-2mm}}
\label{5_global_local_transformer_block}
\end{figure}

It comprises a global-local attention (GLA), a Mix-FFN, and two batch normalizations, as illustrated in Fig. \ref{5_global_local_transformer_block}$\left(a \right)$. Inspired by \cite{wang2022unetformer}, we construct the GLA module to maintain semantically rich global-local contexts in the decoder network. It consists of global and local parallel branches to retain the high-level and fine-grained contexts, as described in Fig. \ref{5_global_local_transformer_block}$\left(b \right)$. \vspace{-3mm}

The local branch has three parallel convolutional layers with $1\times 1$, $3\times 3$, and $5\times 5$ kernel sizes. Different kernel sizes help to capture low-level details of varying-scale objects.
The global branch uses a window-based multi-head self-attention \cite{wang2022unetformer} mechanism to maintain global context.
The output of the global and local branches is summed. As shown in the figure, the summed feature is processed via a depthwise separable convolution. Then the Mix-FFN module processes the output using the same procedure discussed in Section \ref{transformer_encoder}. \vspace{-3mm}

\begin{figure}[h]
\centering
\includegraphics[width=0.9\linewidth]{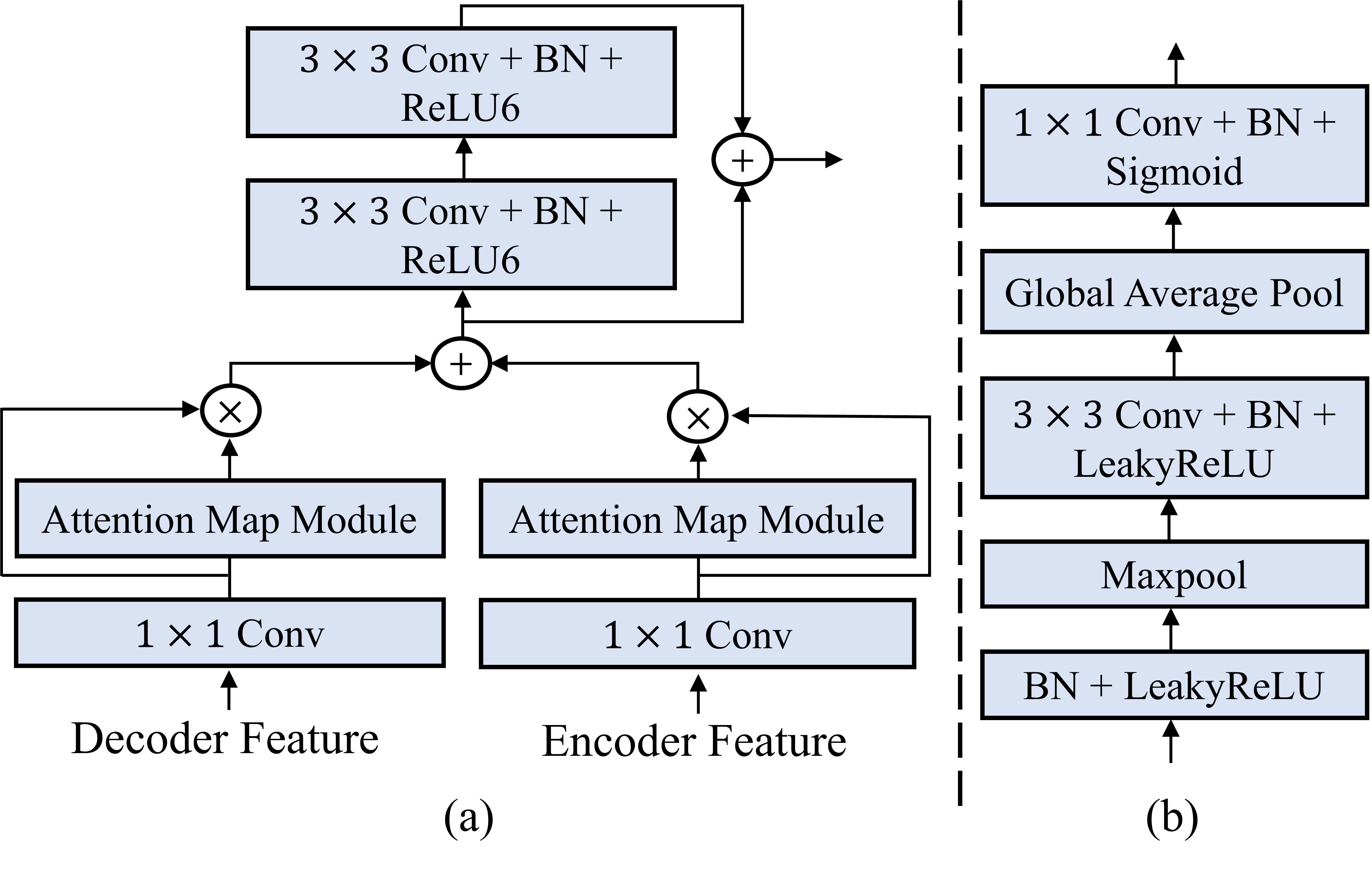}
\caption{(a) The illustration of the fusion module. (b) An attention map module.}
\label{7_fusion_module_new}
\end{figure}

\vspace{-2mm}
\subsubsection{Fusion}
It fuses the GLTB features in the decoder network with their corresponding encoder network features. Figure \ref{7_fusion_module_new} describes its overall design. It comprises two parallel branches that process the encoder and decoder features individually, as shown in Fig. \ref{7_fusion_module_new}$\left(a \right)$.
An AMM takes both features as input. Inspired from \cite{wang2022unetformer}, the AMM is designed to generate a channel-wise attention map $\in c\times 1\times 1$, as displayed in Fig. \ref{7_fusion_module_new}$\left(b \right)$.
After some processing, the input to AMM goes via a global average pooling to obtain an aggregate score for each channel, followed by a sigmoid function.

The channel-wise attention scores are fused with the $1\times 1$ convolution output, as shown in Fig. \ref{7_fusion_module_new}$\left(a \right)$. This fusion enhances the feature representations. The enhanced encoder and the decoder features are now fused using the sum operation.
The fusion module further processes the fused output and adopts a skip connection to strengthen the attribute representation.

\subsubsection{Feature Refinement Module (FRM)}
\begin{figure}[h]
\centering
\includegraphics[width=1.0\linewidth]{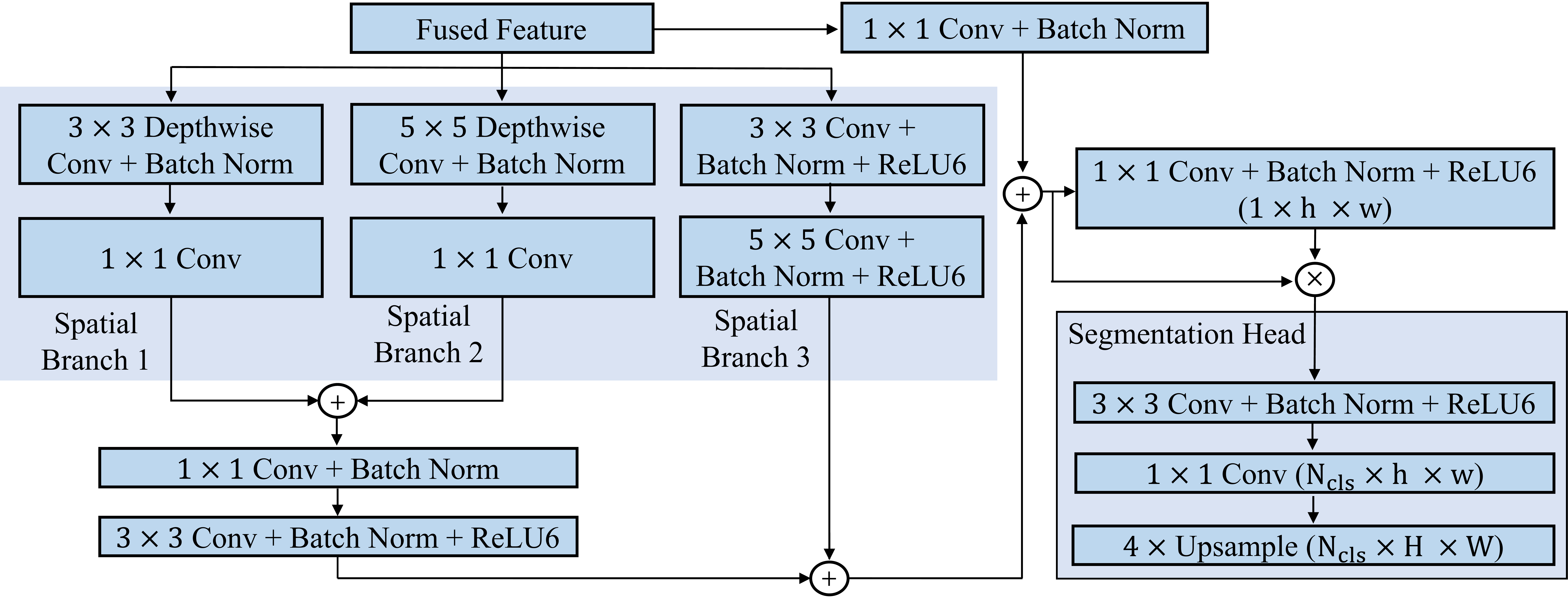}
\caption{The illustration of the feature refinement module. It consists of three parallel spatial branches and a segmentation head.}
\label{8_feature_refinement_module}
\end{figure}

The first stage in the encoder generates low-level spatial features but lacks high-level semantic content. Similarly, the last GLTB in the decoder generates precise semantic details but lacks low-level spatial information. Therefore, this creates a semantic gap between them, and their fused output feature needs refinement. Fig. \ref{8_feature_refinement_module} shows the whole process of feature refinement. \vspace{-3mm}

\vspace{-2mm}
The FRM consists of three parallel spatial branches. The outputs of the first two branches are concatenated, and their dimension is reduced to the decoder channel using a $1\times 1$ convolution.
The fused input feature and the third branch output are aggregated using a sum operation. This feature aggregation helps to enhance the overall representation. Different kernel sizes in the three branches help to learn the representation of multi-scale entities in complex scenes.
The summed output is further processed, followed by a skip connection, and then fed to a segmentation head. It projects the feature channel to the segmentation categories $N_{cls}$ using a $1\times 1$ convolution. At last, it upsamples the feature to the input image size.

\vspace{-5mm}
\section{Experiments \vspace{-2mm}}
\label{sec:sample1}

\subsection{Datasets}
We evaluated the proposed RemoteNet on the LoveDA \cite{wang2021loveda} and Potsdam \cite{potsdam_data} remote sensing segmentation datasets.
The LoveDA dataset contains $5987$ images. These images are split into $2522$ training, $1669$ validation, and $1796$ testing images. It contains seven segmentation categories.
The Potsdam dataset consists of $38$ image tiles. Following previous works \cite{wang2022unetformer, li2021multiattention}, we use $24$ and $14$ tiles for the training and testing, respectively. It contains five foreground semantic labels and a background category, clutter. We ignore the clutter class in the quantitative assessment following the previous works \cite{wang2022unetformer, niu2021hybrid}.

We use the Intersection over Union (IoU), mean Intersection over Union (mIoU), Overall Accuracy (OA), F1 score, and Mean F1 score for the quantitative performance evaluation, following the previous studies \cite{wang2022unetformer, li2021multiattention, niu2021hybrid}.

\vspace{-2mm}
\subsection{Implementation details}
We used the PyTorch library to conduct our experiments on an NVIDIA RTX A6000 GPU. We use cosine annealing as a learning rate scheduler and AdamW as an optimizer with a weight decay of $0.01$. We use a batch size of $8$ for both datasets and cross-entropy as a loss function. Test time augmentation (TTA) techniques are employed during testing.

For the LoveDA dataset, the training uses a random horizontal flip, random scale, and a random crop of $512\times 512$ as data augmentations. We set the base learning rate $6e$$-5$ and trained for the $50$ epochs. We use horizontal flip and multi-scale as the TTA.
We use random scale and random crop of $768\times 768$ as the data augmentations for the Potsdam dataset. We set the base learning rate $6e$$-4$ and trained for the $55$ epochs. During testing, we employ a horizontal flip, vertical flip, ninety-degree rotation, and multi-scale as the TTA.

\vspace{-4mm}
\subsection{Results on LoveDA dataset}
In this subsection, we describe the results of RemoteNet on the LoveDA dataset. Table \ref{lovedatable} shows our quantitative results compared with competing methods. RemoteNet achieved mIoU of 54.56\%, outperforming other competing methods by a decent margin. The precise handling of intrinsic attributes makes it substantially competitive.
RemoteNet significantly improves the background, barren, and agriculture categories compared to other competitive methods. The IoU scores of these classes outperform other competing methods by at least 2.69\%, 0.33\%, and 5.09\%, respectively. IoU of the barren category is comparable with the UNetFormer \cite{wang2022unetformer} and SBSS-MS \cite{cai2023sbss}. However, RemoteNet outperforms other competitive methods significantly in the barren class.
It surpasses the DC-Swin \cite{wang2022novel} and UNetFormer by significant margins in mIoU. RemoteNet beats UNetFormer in the background, road, water, barren, forest, and agriculture classes by 4.30\%, 2.48\%, 0.49\%, 0.33\%, 2.90\%, and 5.66\%, respectively.
It also exceeds by 2.16\% in the mIoU. Similarly, it beats UperNet \cite{wang2022advancing} with the ViTAEv2-S backbone by 1.54\%.

\begin{table}[t]
\centering
\setlength{\tabcolsep}{3pt}
\caption{A quantitative comparison of LoveDA test dataset results with competing methods. The bold and underlined values represent the best and second best scores, respectively.}
\resizebox{\columnwidth}{!}{%
\begin{tabular}{cccccccccc}
\hline
\multirow{2}{*}{Methods} & \multirow{2}{*}{Backbone} & \multicolumn{7}{c}{Class IoU (\%)}                                    & \multirow{2}{*}{mIoU (\%)} \\ \cline{3-9}
                         &                           & Background & Building & Road  & Water & Barren & Forest & Agriculture &                         \\ \hline
FCN8S \cite{long2015fully}  & VGG16                     & 42.60      & 49.51    & 48.05 & 73.09 & 11.84  & 43.49  & 58.30       & 46.69                   \\
FarSeg \cite{zheng2020foreground}    & ResNet50                  & 43.09      & 51.48    & 53.85 & 76.61 & 9.78   & 43.33  & 58.90       & 48.15                   \\
FactSeg \cite{ma2021factseg} & ResNet50                  & 42.60      & 53.63    & 52.79 & 76.94 & 16.20  & 42.92  & 57.50       & 48.94                   \\ 
Semantic-FPN \cite{kirillov2019panoptic}  & ResNet50                  & 42.93      & 51.53    & 53.43 & 74.67 & 11.21  & 44.62  & 58.68       & 48.15                   \\
BANet \cite{wang2021transformer}  & ResT-Lite                 & 43.70      & 51.50    & 51.10 & 79.90 & 16.60  & 44.90  & 62.50       & 49.60                   \\ 
Segmenter \cite{strudel2021segmenter}   & ViT-Tiny                  & 38.00      & 50.70    & 48.70 & 77.40 & 13.30  & 43.50  & 58.20       & 47.10                   \\ 
DC-Swin \cite{wang2022novel}  & Swin-Tiny                 & 41.30      & 54.50    & 56.20 & 78.10 & 14.50  & 47.20  & 62.40       & 50.60                   \\ 
UNetFormer \cite{wang2022unetformer}  & ResNet18                  & 44.70      & \underline{58.80}    & 54.90 & 79.60 & \underline{20.10}  & 46.00  & 62.50       & 52.40                   \\ 
UperNet \cite{wang2022advancing}    & ViTAE-B   + RVSA          & -          & -        & -     & -     & -      & -      & -           & 52.44                   \\ 
UperNet \cite{wang2022advancing}  & ViTAEv2-S                 & -          & -        & -     & -     & -      & -      & -           & 53.02                   \\ 
C-PNet \cite{sun2023target}  & -   & 44.00      & 55.20    & 55.30 & 78.80 & 16.00  & 46.40  & 58.00       & 51.80                   \\
SBSS-MS \cite{cai2023sbss}  & ConvNeXt-T  & \underline{46.31}      & \textbf{62.35}    & \textbf{58.66} & \textbf{82.06} & 19.59  & \textbf{49.48}  & \underline{63.07}     & \underline{54.50}  \\ \hline
RemoteNet (Ours) & MiT-B2  & \textbf{49.00}  & 57.94  & \underline{57.38} & \underline{80.09} & \textbf{20.43}  & \underline{48.90}  & \textbf{68.16}  & \textbf{54.56} \\ \hline
\end{tabular}%
}
\label{lovedatable}
\end{table}

RemoteNet shows the second-highest IoU of road, water, and forest categories after the SBSS-MS. However, the IoU of the background, barren, and agricultural land categories exceeds the SBSS-MS by decent margins of 2.69\%, 0.84\%, and 5.09\%, respectively. It even exceeds its mIoU showing competitive performance.

\begin{figure}[h]
\centering
\includegraphics[width=0.85\linewidth]{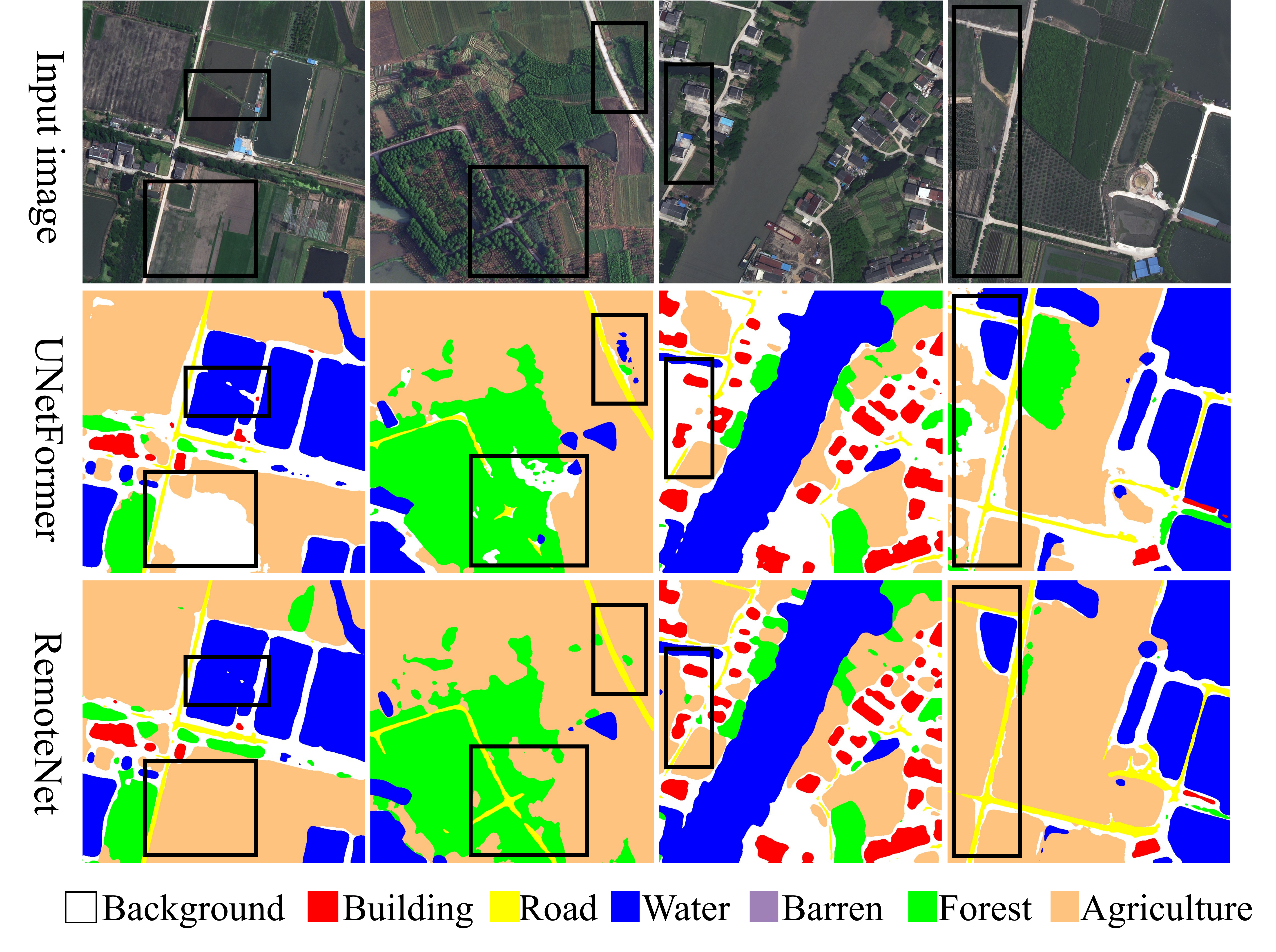}
\caption{The qualitative prediction results of the RemoteNet compared with UNetFormer on the LoveDA test dataset. The regions mark with black boxes show the better prediction of the RemoteNet.}
\label{12_loveda_pred_results_2}
\end{figure}

Next, we illustrate our qualitative prediction results compared with the UNetFormer \cite{wang2022unetformer} in Fig. \ref{12_loveda_pred_results_2}. The proposed RemoteNet generates better segmentation outputs as compared to it. It preserves information about the neighboring pixels in the same category precisely. The UNetFormer struggles to maintain the local details in cluttered scenes. However, our RemoteNet diligently maintains the local context for the water and agricultural land categories in the first column of the figure. It also decently sustains fine-grained information about other classes.
RemoteNet even retains the global context of different classes. Specifically, this is depicted in the third and fourth columns, where it decently captured the global details about the buildings and water classes, respectively. 

\vspace{-3mm}

\vspace{-7.5mm}
\subsection{Results on Potsdam dataset}
In this subsection, we present our experimental findings on the Potsdam dataset. Table \ref{potsdamtable} shows the quantitative comparison results of the RemoteNet with other competing methods.
It shows a mean F1 of 93.27\%, OA of 92.12\%, and mIoU of 87.60\%. It resembles competitive performance against most of the methods.
RemoteNet shows a competitive mean F1 with the MFNet \cite{su2022semantic}, DC-Swin \cite{wang2022novel}, and FT-UNetFormer \cite{wang2022unetformer}. However, it outperforms them in OA by 0.16\%, 0.12\%, and 0.12\%, respectively. It beats the FT-UNetFormer in impervious surface, building, and tree classes by 0.66\%, 0.24\%, and 0.08\%, respectively. RemoteNet shows significant improvement compared with the UNetFormer \cite{wang2022unetformer}. It outperforms it with a decent margin of 0.47\%, 0.82\%, and 0.80\% in the mean F1, OA, and mIoU scores, respectively. It also beats it in the impervious surface, building, low vegetation, and tree categories by a decent margin of 0.96\%, 0.24\%, 0.51\%, and 0.98\%, respectively.

\begin{table}[t]
\centering
\setlength{\tabcolsep}{3pt}
\caption{A quantitative comparison of Potsdam test dataset results with competing methods. The bold and underlined values represent the best and second best scores, respectively.} 
\resizebox{\linewidth}{!}{%
\begin{tabular}{cccccccccc}
\hline
\multirow{2}{*}{Methods} &
  \multirow{2}{*}{Backbone} &
  \multicolumn{5}{c}{F1  score (\%)} &
  \multirow{2}{*}{MeanF1   (\%)} &
  \multirow{2}{*}{OA   (\%)} &
  \multirow{2}{*}{mIoU   (\%)} \\ \cline{3-7}
                 &            & Impervious surface & Building & Low veg & Tree  & Car   &       &       &       \\ \hline
MANet \cite{li2021multiattention}            & ResNet-50  & 93.40          & 96.96    & 88.32  & 89.36 & 96.48 & 92.90 & 91.32 & 86.95 \\
HMANet \cite{niu2021hybrid}           & ResNet101  & 93.90          & \underline{97.60}    & 88.70  & 89.10 & \textbf{96.80} & 93.20 & 92.20 & 87.30 \\
MFNet \cite{su2022semantic}            & ResNet50   & \underline{94.25}          & 97.52    & 88.42  & 89.43 & \underline{96.62} & 93.25 & 91.96 & \underline{87.57} \\
DC-Swin \cite{wang2022novel}          & Swin-S     & 94.19          & 97.57    & 88.57  & 89.62 & 96.31 & 93.25 & 92.00 & 87.56 \\
BANet \cite{wang2021transformer}            & ResT-Lite  & 93.30          & 96.70    & 87.40  & 89.10 & 96.00 & 92.50 & 91.00 & 86.30 \\
UNetFormer \cite{wang2022unetformer}       & ResNet18   & 93.60          & 97.20    & 87.70  & 88.90 & 96.50 & 92.80 & 91.30 & 86.80 \\
FT-UNetFormer \cite{wang2022unetformer}    & Swin-Base  & 93.90          & 97.20    & 88.80  & \underline{89.80} & 96.60 & \textbf{93.30} & 92.00 & 87.50 \\
AFNet \cite{liu2020afnet}            & ResNet          & 94.20          & 97.20    & \textbf{89.20}  & 89.40 & 95.10 & -     & 92.20 & -     \\
SBANet \cite{li2021multitask}           & -          & 93.80          & \textbf{98.00}    & \underline{89.00}  & 89.50 & 94.70 & -     & \underline{92.80} & -     \\
SUD-Net \cite{xu2022transformer}          & -          & 93.61          & 96.98    & 87.63  & 88.70 & 95.95 & 92.57 & \textbf{92.98} & 86.40  \\ \hline
RemoteNet   (Ours) & MiT-B2     & \textbf{94.56}          & 97.44    & 88.21  & \textbf{89.88} & 96.26 & \underline{93.27} & 92.12 & \textbf{87.60} \\ \hline
\end{tabular}%
}
\label{potsdamtable}
\end{table}

Our F1 score of building and low vegetation are competitive with the AFNet \cite{liu2020afnet} and SBANet \cite{li2021multitask}. However, we outperform AFNet by a good margin of 0.36\%, 0.48\%, and 1.16\% in the impervious surface, tree, and car categories, respectively. Similarly, in those classes, it exceeds the SBANet by a 0.76\%, 0.38\%, and 1.56\%, respectively.
Our OA is competitive with the SUD-Net \cite{xu2022transformer}. However, we outperform it by 0.70\% and 1.20\% margins in the mean F1 and mIoU scores, respectively. It also exceeds the SUD-Net in all classes.

\begin{figure}[h]
\centering
\includegraphics[width=0.85\linewidth]{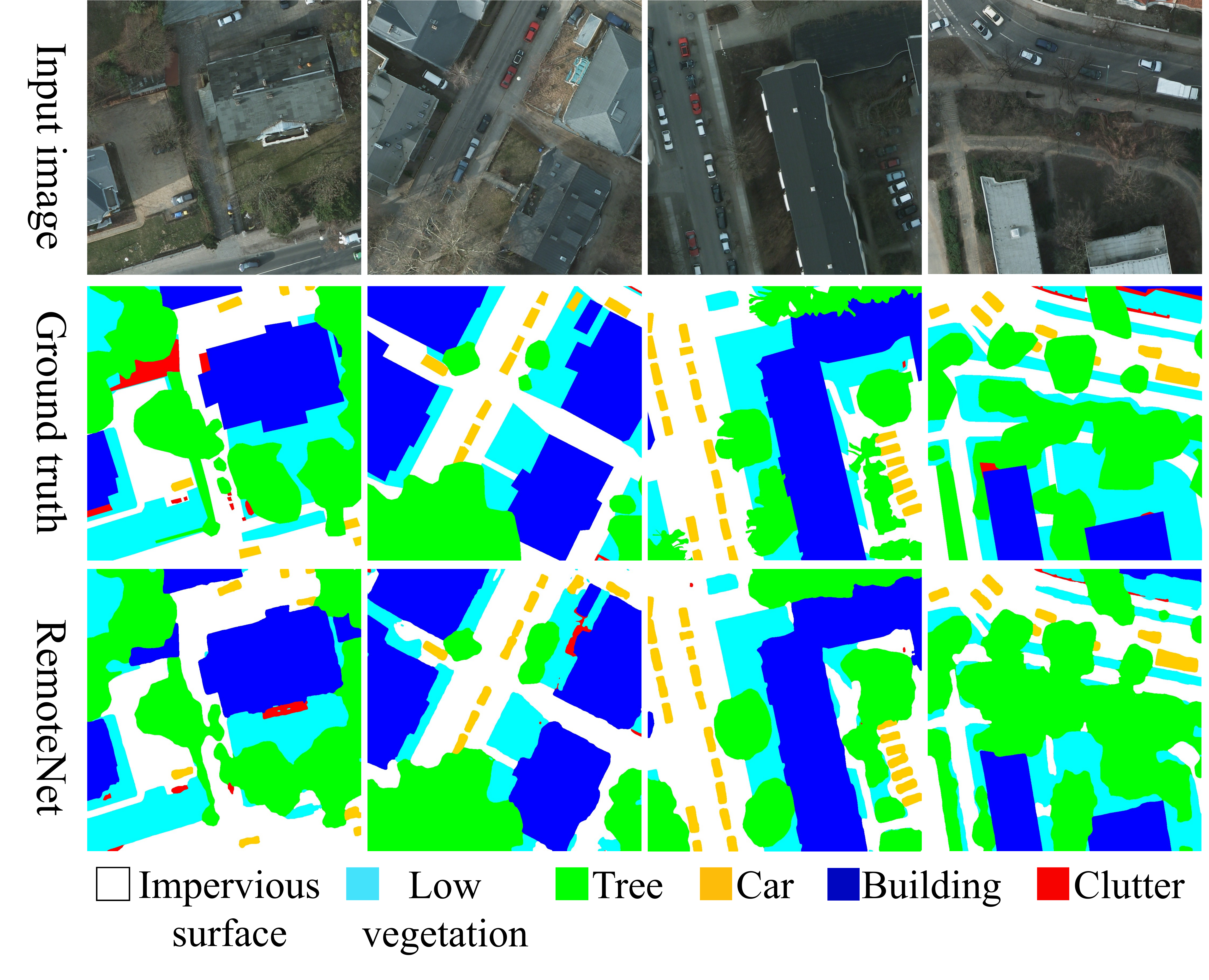}
\caption{The qualitative prediction results of the RemoteNet on the Potsdam test dataset.}
\label{13_potsdam_pred_results_1}
\end{figure}

We present the qualitative results of the RemoteNet on the Potsdam test images in Fig. \ref{13_potsdam_pred_results_1}. It yields precise results while maintaining the global and local context details.
It also maintains the global context about small objects like cars accurately. It preserves its shape and structure diligently.
Overall, the proposed RemoteNet generates smooth segmentation while retaining the inherent information of different category objects in the complex environment. \vspace{-3mm}

\vspace{-7mm}
\subsection{Ablation Study}
We discuss the sensitivity of different modules in this subsection. Table \ref{loveda_ablation} presents the quantitative comparison results on the LoveDA test dataset.
The first row in the table represents the effect on the performance without using the AMM. Removing the AMM shows a significant reduction of 2.26\% in the mIoU.
After that, we test the effect of a number of spatial branches in the FRM. We remove the third branch in the FRM and experiment with the remaining two branches. The second row in the table shows the result with the two branches. Removing the third branch reduces the mIoU by 1.05\%.
Next, we investigate the sensitivity of the whole FRM. The deepest decoder feature is directly sent to the segmentation head for the prediction. The third row in the table shows the result without using the FRM. It reduces the mIoU by 1.85\%.
Then after, we also test the impact of the fusion module. We directly fuse the encoder and decoder features using an element-wise addition without our fusion module. The fourth row in the table shows the performance without the fusion module. Removing this module shows a significant reduction of 1.78\% in the mIoU.
The IoU score of the building is slightly better than ours in all the sensitivity tests. However, these gains come with significantly reducing the IoU of other classes. The proposed RemoteNet outperforms them significantly in other categories.
\vspace{-3.5mm}

\begin{table}[t]
\centering
\setlength{\tabcolsep}{3pt}
\caption{Ablation Experimental Results on the LoveDA test datasets.}
\resizebox{\columnwidth}{!}{%
\begin{tabular}{ccccccccc}
\hline
\multirow{2}{*}{Methods}    & \multicolumn{7}{c}{Class   IoU (\%)}                                  & \multirow{2}{*}{mIoU   (\%)} \\ \cline{2-8}
                            & Background & Building & Road  & Water & Barren & Forest & Agriculture &                              \\ \hline
Without   AMM               & 45.22      & 59.25    & 55.67 & 77.83 & 14.88  & 48.75  & 64.50       & 52.30                        \\
Two spatial branches in FRM & 47.29      & \textbf{59.81}    & 55.45 & 79.55 & 18.07  & 47.81  & 66.59       & 53.51                        \\
Without   FRM               & 46.23      & 58.64    & 56.73 & 79.87 & 14.37  & 46.67  & 66.48       & 52.71                        \\
Without   fusion   module   & 46.34      & 58.56    & \textbf{57.42} & 79.14 & 14.22  & 48.20  & 65.56       & 52.78                        \\ \hline
RemoteNet (Ours)            & \textbf{49.00}      & 57.94    & 57.38 & \textbf{80.09} & \textbf{20.43}  & \textbf{48.90}  & \textbf{68.16}       & \textbf{54.56}                        \\ \hline
\end{tabular}%
}
\label{loveda_ablation}
\end{table}

\section{Conclusion}
In this letter, we introduce RemoteNet, a remote sensing image segmentation Network, designed explicitly for semantic segmentation of remote sensing images. RemoteNet demonstrates a significant advantage in preserving both global and fine-grained local context information in complex scenes. The constructed decoder network generates precise segmentation output, ensuring accurate delineation of objects. The integration of multi-scale features enables the capture of contextual details across varying complexities, further enhancing the segmentation performance. Through a series of ablation studies, we demonstrate the significance of the proposed modules. Experimental findings on the Potsdam and LoveDA segmentation datasets showcase the clear advantage of RemoteNet, highlighting its competitive performance and effectiveness. Future work will focus on developing a more robust architecture to capture complex traits for better segmentation results.
\vspace{-3.5mm}

\section*{Acknowledgment}
This work was supported by the National Research Foundation of Korea (NRF) grant funded by the Korean Government (MSIT) (No. 2021R1C1C1012590), (No. 2022R1A4A1023248) and the Information Technology Research Center (ITRC) support program supervised by the Institute of Information Communications \& Technology Planning \& Evaluation (IITP) grant funded by the Korean Government (MSIT) (IITP-2023-2020-0-01808).
\vspace{-4.5mm}

\bibliographystyle{IEEEtran}
\bibliography{my_bib}

\end{document}